# FIVE MODULUS METHOD FOR IMAGE COMPRESSION

Firas A. Jassim , Hind E. Qassim

Management Information System Department, Irbid National University, Irbid-Jordan
Firasajil@yahoo.com
Management Information System Department, Toledo College, Irbid-Jordan
Hindeq@yahoo.com

## ABSTRACT

*Data may be compressed by reducing the redundancy in the original data, but this makes the data have more errors. In this paper a novel approach of image compression based on a new method that has been created for image compression which is called Five Modulus Method (FMM). The new method consists of converting each pixel value in an 8×8 block into a multiple of 5 for each of the R, G and B arrays. After that, the new values could be divided by 5 to get new values which are 6-bit length for each pixel and it is less in storage space than the original value which is 8-bits. Also, a new protocol for compression of the new values as a stream of bits has been presented that gives the opportunity to store and transfer the new compressed image easily.*

## 1. INTRODUCTION

Image compression methods aim at representing an approximation of original images with as few bits as possible while controlling the quality of these representations. Now-a-days, image compression techniques are very common in a wide area of researches. Two types of image compression have been introduces which are lossless and lossy compression. With lossless compression, the ability to reconstruct the original image after compression is exact. It is rear to obtain error-free compression more than 2:1 ratio. On the other hand, in lossy compression the ratio can be obtained with some error between the original image and the reconstructed image. In many cases, error-free reconstruction of the original image may be impossible. If the image has some noise, then there are denoising methods that can be used to reduce that noise. Therefore, lossy compression may produce an acceptable error that does not affect too much to the original image. This can be seen in fast transmission of still images over the Internet where the amount of error can be acceptable [11] [12].

The attainable ratio for lossless compression coders is 2:1 for most of the images. Although, lossy algorithms produce some error, but it may give high compression ratios. The recommendation ISO DIS 10 918-1 known as JPEG Joint Photographic Experts Group [10]. JPEG has become an international standard for lossy compression of still image. Image quality is measured using peak signal-to-noise ratio (PSNR) [1] as most common objective measure. Whenever the value of PSNR is high this implies good compression because it means high Signal to Noise. In other words, the signal represents the original image while the noise represents the error in reconstruction. So, a compression scheme having a high PSNR can be recognized as a better one [6]. Therefore, certain performance measures have been established to compare different compression algorithms. Compression Ratio (CR) [2] is defined as the ratio of the number of bits required to represent the data before compression to the number of bits required after compression.

## 2. DIGITAL IMAGE BACKGROUND

A digital image is a rectangular array of dots, or pixels, arranged in m rows and n columns. A digital image is represented by a two-dimensional array of pixels, which are arranged in rows and columns. Hence, a digital image can be presented as M×N array [7].

$$f(x,y) = \begin{bmatrix} f(0,0) & f(0,1) & \cdots & f(0,N-1) \\ f(1,0) & f(1,1) & \cdots & f(1,N-1) \\ \vdots & \vdots & \ddots & \vdots \\ f(M-1,0) & f(M-1,1) & \cdots & f(M-1,N-1) \end{bmatrix}$$

where f(0,0) gives the pixel of the left top corner of the array that represents the image and f(M-1,N-1) represents the right bottom corner of the array. A Grey-scale image, also referred to as a monochrome image contains the values ranging from 0 to 255, where 0 is black, 255 is white and values in between are shades of grey. In color images, each pixel of the array is constructed by combining three different channels (RGB) which are R=red, G=green and B=blue. Each channel represents a value from 0 to 255. In digital image, each pixel is stored in three bytes, while in a Grey image is represented by only one byte. Therefore, color images take three times the size of Gray images.

## 3. IMAGE COMPRESSION

The main idea in image compression is to reduce the data stored in the original image to a smaller amount. According to scientific revolution in the internet and the expansion of multimedia applications, the requirements of new technologies have been grown. Recently, many different techniques have been developed to address these requirements for both lossy and lossless compression [2][4]. Modern computers employ graphics extensively. Window-based operating systems display the disk's file directory graphically. The progress of many system operations, such as downloading a file, may also be displayed graphically. Now-a-days, most of the applications working under windows provide a graphical user interface (GUI), which makes it easier to use the program. Many areas of life use computer graphics to change the type of the problem from information to a digital image. Thus, images are important, but they tend to be big! Modern hardware can display many colors, which is why it is common to have a pixel represented internally as a 24-bit number, where the percentages of red, green, and blue occupy 8 bits each. Such a 24-bit pixel can specify 16.78 million colors. As a result, an image at a resolution of 512×512 that consists of such pixels occupies 786,432 bytes. At a resolution of 1280×800 it gives 3,072,000 bytes as a total for 3 bytes for each pixel which makes it four times as big as 512×512 [2]. Therefore, an image compression came arise so solve this problem. In this article, we have focused our attention on lossy compression using Five Modulus Method that can be seen in the next section.

# 4. FIVE MODULUS METHOD

In most of images, there is a common feature which is the neighboring pixels are correlated. Therefore, finding a less correlated representation of image is one of the most important tasks. One of the basic concepts in compression is the reduction of redundancy and Irrelevancy. This can be done by removing duplication from the image. Sometime, Human Visual System (HVS) can not notice some parts of the signal, i.e. omitting these parts will not be noticed by the receiver. This is called as Irrelevancy.

Also, for bi-level images, the principle of image compression tells us that the neighbors of a pixel tend to be similar to the pixel. According to [2], this principle can be extended as that if the current pixel has any color (black or white), then pixels seen in the past or future of the same color tend to have the same neighbors.

Hence, our proposed technique which is called Five Modulus Method (shortly FFM) is consists of dividing the image into blocks of 8×8 pixels each. Clearly, we know that each pixel is a number between 0 to 255 for each of the Red, Green, and Blue arrays. Therefore, if we can transform each number in that range into a number divisible by 5, then this will not affect the Human Visual System (HVS). Mathematically speaking, any number divided by 5 will give a remainder ranges from 0-4 (e.g., 15 mod 5 is 0, 17 mod 5 is 2, 201 mod 5 is 1, 187 mod 5 is 2 and so on). Here, we have proposed a new formula to transform any number in the range 0-255 into a number that when divided by 5 the result is always lying between 0-4.

Therefore, the pixels 200, 201, and 202 are the same for the human eye. Hence, a novel algorithm have been proposed to transform each pixel in the range 0-255 into the following numbers 0,5,10,15,20,25,30,35,40,…,200, 205,210,215,...,250, 255, (i.e. multiples of 5). Actually, any number in the range 0-4 (which is the remainder of dividing 0-255 by 5) can be transformed as follows 0→(same pixel), 1→(-1), 2→(-2), 3→(+2), 4→(+1). The algorithm can be described as:

## FMM Algorithm

The basic idea in FMM is to check the whole pixels in the 8×8 and transform each pixel into a number divisible by 5 according to the following conditions.

```
if A(i,j) Mod 5 = 4
        A(i,j)=A(i,j)+1
Else if A(i,j) Mod 5 = 3
        A(i,j)=A(i,j)+2
Else if A(i,j) Mod 5 = 2
        A(i,j)=A(i,j)-2
Else if A(i,j) Mod 5 = 1
        A(i,j)=A(i,j)-1
```

Where A(i,j) is the digital image representation of the 8×8 block for any of the R,G, or B arrays consisting the digital image. The following table shows the transformation exactly,

Table 1. New values obtained by FMM

| Old | New | | Old | New |
|-----|-----|-----|-----|-----|
| 0 | 0 | | 100 | 100 |
| 1 | 0 | | 101 | 100 |
| 2 | 0 | | 102 | 100 |
| 3 | 5 | | 103 | 105 |
| 4 | 5 | | 104 | 105 |
| 5 | 5 | | 105 | 105 |
| 6 | 5 | … | 106 | 105 |
| 7 | 5 | | 107 | 105 |
| 8 | 10 | | 108 | 110 |
| 9 | 10 | | 109 | 110 |
| 10 | 10 | | 110 | 110 |
| 11 | 10 | | 111 | 110 |
| 12 | 10 | | 112 | 110 |
| 13 | 15 | | 113 | 115 |
| 14 | 15 | | 114 | 115 |
| 15 | 15 | | 115 | 115 |

In addition, we can see that the new pixels are always having zero remainder when divided by 5. Consequently, the resulting numbers are multiples of 5 between 0-255, which are 52 numbers (0,5,10,15,20,…,255). Hence, if we divide these numbers by 5 again we will get remainder range from 0-51. Now, let us take a practical example, the following table is an exactly 8×8 block have been taken from any of the Red, Yellow, or Green arrays in any arbitrary BMP image,

Table 2. Original 8×8 block as a sample

| 221 | 232 | 231 | 242 | 246 | 247 | 251 | 250 |
|-----|-----|-----|-----|-----|-----|-----|-----|
| 220 | 227 | 231 | 236 | 242 | 241 | 250 | 251 |
| 221 | 215 | 221 | 232 | 240 | 247 | 251 | 251 |
| 217 | 216 | 216 | 225 | 237 | 241 | 245 | 247 |
| 216 | 221 | 217 | 222 | 231 | 235 | 242 | 247 |
| 220 | 216 | 222 | 215 | 227 | 231 | 242 | 247 |
| 216 | 216 | 211 | 216 | 222 | 227 | 237 | 247 |
| 217 | 216 | 211 | 216 | 217 | 222 | 237 | 235 |

Now, to measure the dispersion between pixels [8], we can calculate the standard deviation in Table (2) as (12.8285). After FFM is applied, i.e. transform each pixel into multiple of 5, the new 8×8 block can be seen in the following table,

Table 3. Transformed 8×8 block using FMM

| 220 | 230 | 230 | 240 | 245 | 245 | 250 | 250 |
|---|---|---|---|---|---|---|---|
| 220 | 225 | 230 | 235 | 240 | 245 | 250 | 250 |
| 220 | 215 | 220 | 230 | 240 | 245 | 250 | 250 |
| 215 | 215 | 215 | 225 | 235 | 240 | 245 | 245 |
| 215 | 220 | 215 | 220 | 230 | 235 | 240 | 245 |
| 220 | 215 | 220 | 215 | 225 | 230 | 240 | 245 |
| 215 | 215 | 210 | 215 | 220 | 225 | 235 | 245 |
| 215 | 215 | 210 | 215 | 215 | 220 | 235 | 235 |

Dividing the new block by 5, we get

Table 4. Dividing FMM block by 5

| 44 | 46 | 46 | 48 | 49 | 49 | 50 | 50 |
|---|---|---|---|---|---|---|---|
| 44 | 45 | 46 | 47 | 48 | 49 | 50 | 50 |
| 44 | 43 | 44 | 46 | 48 | 49 | 50 | 50 |
| 43 | 43 | 43 | 45 | 47 | 48 | 49 | 49 |
| 43 | 44 | 43 | 44 | 46 | 47 | 48 | 49 |
| 44 | 43 | 44 | 43 | 45 | 46 | 48 | 49 |
| 43 | 43 | 42 | 43 | 44 | 45 | 47 | 49 |
| 43 | 43 | 42 | 43 | 43 | 44 | 47 | 47 |

Now, to measure the dispersion between pixels, we can calculate the standard deviation in Table (4) as (2.572751). We can see that the standard deviation in the original block is higher than the transformed block which means that the dispersion between data becomes less after the transformation using FMM. As a good result, the new block is highly correlated, i.e. the new data are less dispersed compared with the original data because they are close to each other which will give less dispersion.

After that, the minimum of the new block shown in table (4) can be found and subtracted from the whole block. According to table (4), the minimum is 42 and after subtracting it from the block we can get a new block as:

Table 5. After subtracting minimum

| 2 | 4 | 4 | 6 | 7 | 7 | 8 | 8 |
|---|---|---|---|---|---|---|---|
| 2 | 3 | 4 | 5 | 6 | 7 | 8 | 8 |
| 2 | 1 | 2 | 4 | 6 | 7 | 8 | 8 |
| 1 | 1 | 1 | 3 | 5 | 6 | 7 | 7 |
| 1 | 2 | 1 | 2 | 4 | 5 | 6 | 7 |
| 2 | 1 | 2 | 1 | 3 | 4 | 6 | 7 |
| 1 | 1 | 0 | 1 | 2 | 3 | 5 | 7 |
| 1 | 1 | 0 | 1 | 1 | 2 | 5 | 5 |

As we can see from table (5) that the maximum number is 8 and it's representation in binary coding is (1000) which is four bit length. Hence, the entire block length is 64×4=256, where the original length is 64×8=512. Therefore, the compression ratio is 512/256=2. If we take different block the result may also different which depends on the variation between data in the desired block.

## 5. FIVE MODULUS METHOD BIT STORAGE

Clearly, as we know that each number 0-255 is stored at 8-bits location (1 byte). After transformation into FMM and making each number between 0-51 (255/5=51) and this can be shown in table (6) as:

Table 6. Bit stream representation

| Pixel | Bit stream | | Pixel | Bit stream |
|-------|-----------|---|-------|-----------|
| 0 | 000000 | | 36 | 100100 |
| 1 | 000001 | | 37 | 100101 |
| 2 | 000010 | | 38 | 100110 |
| 3 | 000011 | | 39 | 100111 |
| 4 | 000100 | | 40 | 101000 |
| 5 | 000101 | | 41 | 101001 |
| 6 | 000110 | … | 42 | 101010 |
| 7 | 000111 | | 43 | 101011 |
| 8 | 001000 | | 44 | 101100 |
| 9 | 001001 | | 45 | 101101 |
| 10 | 001010 | | 46 | 101110 |
| 11 | 001011 | | 47 | 101111 |
| 12 | 001100 | | 48 | 110000 |
| 13 | 001101 | | 49 | 110001 |
| 14 | 001110 | | 50 | 110010 |
| 15 | 001111 | | 51 | 110011 |

Clearly, from table (6) we can see that each number 0-51 needs 6-bits representation which is less by two bits from the traditional bit storage for any ASCII coding representation which is 8-bits. Now, we can construct a new protocol to accommodate the new pixel coordinates which will be discussed in the next section.

## 6. FMM PROTOCOL

According to the previous section, we concluded that each new pixel needs 6-bits storage instead of the standard 8-bits storage and using minimum subtraction stated earlier, the new protocol can be established which is similar to that type of protocols used in computer networks [9]. Firstly, for each 8×8 block, the new stream starts with 6 bits in the beginning that is reserved for the value of the minimum for the block after transformation. Secondly, a bit for repetition, i.e. if the whole block consists of the same value, this value will be 1 otherwise its 0. This bit is used to decrease storage for each block when there is no need to repeat values if the entire block consisting of the same value. Since our proposed method transform value into multiples of 5, this will lead to less dispersion in the same block. Mostly, the repetition bit will be 1 for most of the times because of the Rapprochement between block values. Next, the value of the maximum for the current block can be added to determine the width of the current stream values. Finally, a stream of {max length}-bits, i.e. the length of the maximum will be the standard length for all other values in that stream, for each new pixel coordinate will follow. Practically, we will illustrate this by the following example, suppose a similar 8×8 block from an arbitrary image. After transformation into FMM, we have found that each pixel in the 8×8 block have the same value as:

Table 7. Similar 8×8 block representation

| 11 | 11 | 11 | 11 | 11 | 11 | 11 | 11 |
|----|----|----|----|----|----|----|----|
| 11 | 11 | 11 | 11 | 11 | 11 | 11 | 11 |
| 11 | 11 | 11 | 11 | 11 | 11 | 11 | 11 |
| 11 | 11 | 11 | 11 | 11 | 11 | 11 | 11 |
| 11 | 11 | 11 | 11 | 11 | 11 | 11 | 11 |
| 11 | 11 | 11 | 11 | 11 | 11 | 11 | 11 |
| 11 | 11 | 11 | 11 | 11 | 11 | 11 | 11 |
| 11 | 11 | 11 | 11 | 11 | 11 | 11 | 11 |

The FFM protocol can be represented as

| 001011 | 1 |
|--------|---|

Were (001011) is the representation of (11), taken with 6-bits length, and the bit after is for the repetition bit which is 1, this mean that the whole block is consisting of the same value but what about not similar values. According to table (4), we can see that the maximum is (7) which is (111) in binary. Therefore, max length in the stream is 3 and the FFM protocol for the block given in table (2) can be represented as:

| 101010 | 0 | 000111 | 010 | 100 | …. | 101 |
|--------|---|--------|-----|-----|-----|-----|

Where (101010) is the representation for the minimum of the block which is (42), see table (4), and the repetition bit is set to zero because the values in the block are not similar to each other. Next, the maximum number of the resulted block after subtracting is (7), see table (5). The shaded cells are facing the shaded cells shown in table (5). Hence, the general bit stream for the FMM protocol can be represented by the following figure.

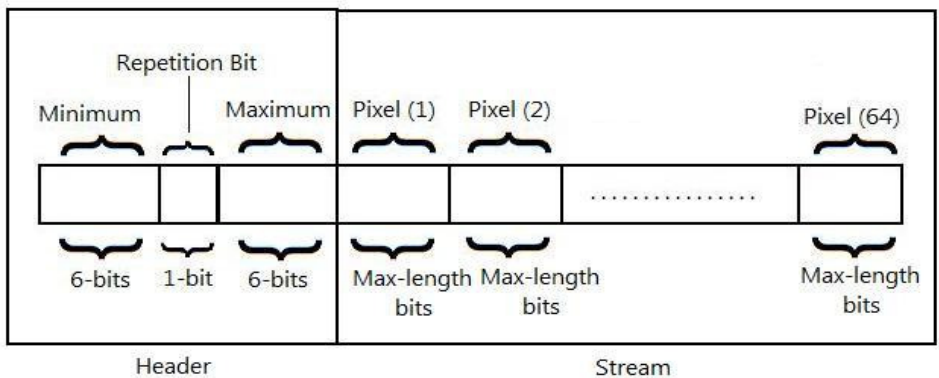

Figure 1. FMM stream protocol

# 7. ERROR MEASURES

In order to measure the quality of the reconstructed images compared with the original ones, a standard metric to measure this quality have been established. The most usable metric by image coders is the peak signal to noise ratio (PSNR). Practically, let us denote the pixels of the original image by Pi and the pixels of the reconstructed image by Qi (where $1 \leq i \leq n$), we first define the mean square error (MSE) [7] between the two images as:

$$MSE = \frac{1}{n} \sum_{i=1}^{n} (P_i - Q_i)^2$$

It is the mathematical mean of the differences in the values for the pixels between the original and the reconstructed images. The root mean square error (RMSE) is defined as the square root of the MSE [5]:

$$RMSE = \sqrt{\frac{1}{n} \sum_{i=1}^{n} (P_i - Q_i)^2}$$

Hence, the PSNR can be defined as:

$$PSNR = 20 \, Log_{10} \, \frac{\max |P_i|}{RMSE}$$

Also, the compression ratio may be defined as the ratio between the reconstructed image to the original image. Therefore, suppose P and Q are two units of a set of data representing the same information. The compression ratio, CR, is denoted as [7],

$$CR = \frac{Q_i}{P_i}$$

# 8. EXPERIMENTAL RESULTS

Actually, our proposed FMM algorithm is not an optimum or even near optimum compression technique. Experimentally, four types of standard test images have been used which are : Lena, Baboon, Peppers, and F16, see Figure (2). Actually, FFM was implemented to the four test images and the result were as follows,

Table 8. PSNR for test images

|        | PNSR    |         |
|--------|---------|---------|
|        | JPG     | FMM     |
| Lena   | 33.2093 | 44.376  |
| Baboon | 26.2079 | 53.2149 |
| Pepper | 30.286  | 47.96   |
| F16    | 32.6075 | 40.0879 |

Table 9. CR for test images

| | CR | |
|---|---|---|
| | JPG | FMM |
| Lena | 20.81 | 1.61 |
| Baboon | 10.16 | 1.39 |
| Pepper | 18.96 | 1.57 |
| F16 | 20.26 | 1.87 |

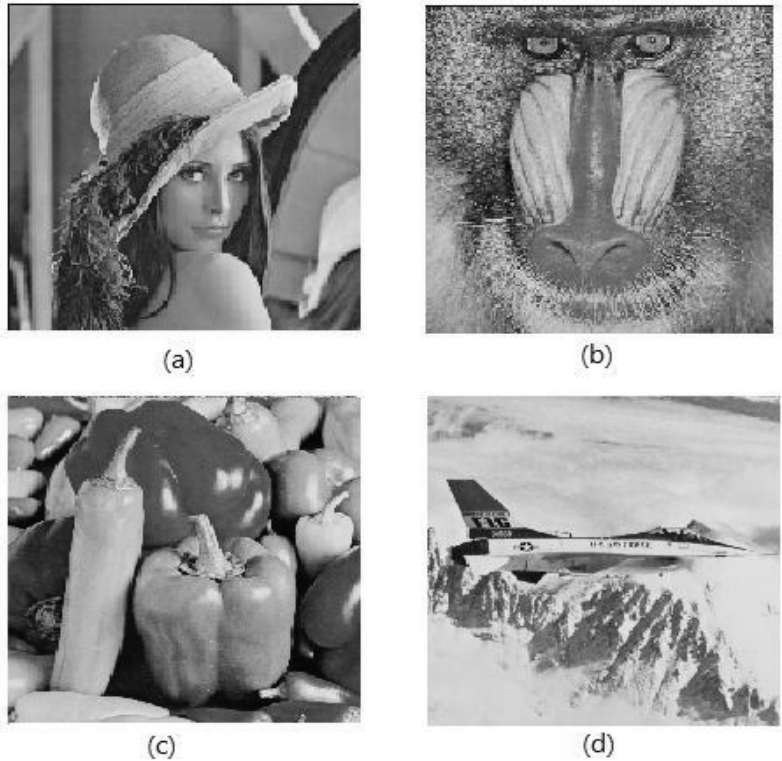

Figure 2. Four test images (a) Lena (b) Baboon (c) Peppers (d) F16

## 9. CONCLUSIONS

Clearly, our proposed method is not the optimal compression technique but it can help to improve some existing compression methods. This paper demonstrates the potential of the FMM based image compression technique. The advantage of this method is the high PSNR although it's low compression ratio. This method is appropriate for bi-level images (black and white medical images) where the pixel in such images is represented by one byte (8-bit). The proposed method can not be used as a standalone method because of it's low compression ratio but it can be used as a scheme embedded into other compression techniques (such as JPEG) to reduce compression ratio. As a recommendation, a Variable Modulus Method (X) MM, where X can be any number, may be constructed in later research.